\pdfoutput=1

\documentclass[11pt]{article}

\usepackage[]{EMNLP2022}

\usepackage{times}
\usepackage{latexsym}
\usepackage{caption,subcaption}
\usepackage{algorithmic}
\usepackage[ruled,vlined]{algorithm2e}
\usepackage{subcaption}
\usepackage{amssymb}
\usepackage{multirow}
\usepackage{longtable}
\usepackage{booktabs}
\usepackage{graphicx}
\usepackage{mathtools}
\usepackage{geometry}
\usepackage{xcolor}
\usepackage{pifont}
\usepackage{tabularx}
\usepackage{comment}
\usepackage{cleveref}
\usepackage[normalem]{ulem}

\usepackage[T1]{fontenc}

\usepackage[utf8]{inputenc}

\usepackage{microtype}

\usepackage{inconsolata}

\title{PIE-QG: Paraphrased Information Extraction for Unsupervised Question Generation from Small Corpora}

\author{Dinesh Nagumothu, Bahadorreza Ofoghi, Guangyan Huang, Peter W. Eklund  \\ 
  School of Information Technology, Deakin University, 221 Burwood Hwy, Burwood 3125\\ Victoria, Australia\\
  \texttt{\small\{dnagumot,b.ofoghi,guangyan.huang,peter.eklund\}@deakin.edu.au}}
  
\begin{document}

\maketitle
\begin{abstract}
Supervised Question Answering systems (QA systems)  
rely on domain-specific human-labeled data for training.  Unsupervised QA systems generate their own question-answer training pairs, typically using  secondary knowledge sources to achieve this outcome. 
Our approach (called PIE-QG) uses Open Information Extraction (OpenIE) to generate synthetic training questions from paraphrased passages and uses the question-answer pairs as training data for a language model for a state-of-the-art QA system based on BERT. Triples in the form of <subject, predicate, object> are extracted from each passage, and questions are formed with subjects (or objects) and predicates while objects (or subjects) are considered as answers. Experimenting on five extractive QA datasets demonstrates that our technique achieves on-par performance with existing state-of-the-art QA systems with the  benefit of being trained on an order of magnitude fewer documents and without any recourse to external reference data sources. 
\end{abstract}

\section{Introduction}
Question Answering systems (QA systems) provide  answers to input questions posed in natural language. Answering questions from unstructured text can be performed using Machine Reading Comprehension (MRC). Given a passage, several sentences or a paragraph, and a question posed, the QA system produces the best suitable answer. Extractive Question Answering  systems (EQA systems) are a subset of QA systems and involve an MRC task where the predicted answer is a span of words from the passage. With pre-trained language models~\cite{radford2018improving}, EQA systems
 achieve excellent results, surpassing even human performance. Pre-trained language models, such as BERT~\cite{devlin-etal-2019-bert} and GPT~\cite{radford2019language}, can be fine-tuned to perform downstream tasks such as QA. However, huge amounts of data are required to train these models for specific domains, making the task labor-intensive, in terms of the effort required to assemble suitable domain-specific training data.

A single training instance for an EQA system dataset requires a question, a passage, and an answer. Domain-relevant documents can be collected with advanced information retrieval tools, and passages are formed by splitting documents into several related sentences or a paragraph.
However, generating the question and answer pairs, that provide the training set for the QA system from a given passage, is considered the most difficult challenge, an approach known as unsupervised QA~\cite{cui2004unsupervised}.

\begin{figure*}[t!]
    \centering
    \includegraphics[width=\textwidth]{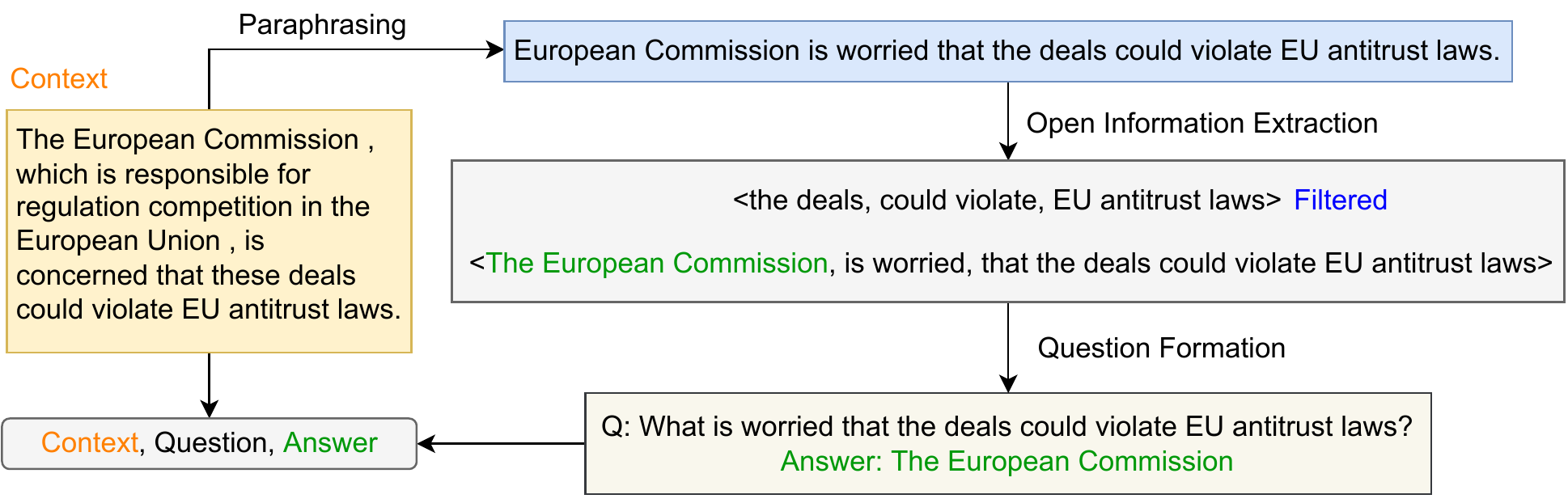}
    \caption{Question Generation from a context (left) by paraphrasing followed by information extraction using OpenIE. Note: The text in green indicates the selected answer.}
    \label{fig:qa-sample}
\end{figure*}

Existing unsupervised QA system techniques such as \cite{lewis-etal-2019-unsupervised} and \cite{lyu-etal-2021-improving} use an out-of-domain dataset for question generation, namely, they require additional training sources beyond what can be provided by the target corpus and a pre-trained generic model. 
On the other hand, rule-based QA system methods, those constrained to generate question-answer pairs from only the corpus itself, run the risk of generating questions with high lexical overlap with the passage, at risk of forcing the model to learn word matching patterns. The work of  \cite{fabbri-etal-2020-template} and \cite{li-etal-2020-harvesting} use information retrieval-based methods, such as elastic search and citation navigation, to create questions from passages other than those presented within the target dataset. However, these methods may not generate sufficient training questions, especially when the corpus is small and has no citation or inter-document linking structure.

In this paper, we focus on addressing the limitations of EQA systems using a novel unsupervised Paraphrased Information Extraction for Question Generation (PIE-QG) method that generates synthetic training data through the extraction of <subject, relation, object> triples from a given corpus. We use the original passage to produce question-answer training pairs by generating a paraphrased version of the original passage to avoid lexical overlap between the passage and the question-answers. We adopt Open Information Extraction~\cite{kolluru-etal-2020-openie6} to extract <subject, relation, object> triples from every sentence of the paraphrased passage. These triples 
are rich in semantics  and represent raw facts; therefore, generating question-answer pairs from triples results in well-formed and effective training data. Furthermore, many sentences in the passage contribute to generating meaningful extractions, thus helping to pose questions in different ways from a single passage. An  example of the question generation process we propose (called PIE-QG for Paraphrasing, Information Extraction Question Generation) is shown in Figure~\ref{fig:qa-sample}. 
The contributions of this paper are as follows:

\begin{enumerate}
    \item We describe the PIE-QG method in which paraphrased passages from the original corpus are used to generate question-answer pairs without  reliance on external reference data sources, such as retrieval-based or inter-document link navigation methods.  Paraphrasing passages  reduces the effect of lexical overlap between the passage and the question.
    \item We generate multiple questions from a single paraphrased passage by adopting Open Information Extraction to extract facts, thus increasing the number of question-answer pairs extracted from the corpus. 
\end{enumerate}

We have conducted experiments on four Extractive QA datasets and demonstrate that the proposed PIE-QG method achieves comparable performance in terms of Exact Match (EM) and F1 score while requiring significantly fewer passages.

The remainder of this paper is organized as follows. We present related work in Section \ref{sec:related}. In Section \ref{sec:question_generation}, we describe the proposed PIE-QG method. Section \ref{sec:experiments} discusses the experimental setup. In Section \ref{sec:results}, we evaluate the performance of our method. Section \ref{sec:limitations} presents the limitations of the proposed method and Section \ref{sec:conclusion} offers some concluding remarks.

\section{Related Work}
\label{sec:related}
Pre-trained language models, such as BERT~\cite{devlin-etal-2019-bert}, can be fine-tuned for downstream tasks like Extractive QA systems (EQA systems). A comprehensive natural language (NL) passage, which might be several sentences or a paragraph of NL-text, is considered as the context where the model finds the answer span. The input question and the context are represented as a single sequence, passed to a pre-trained model and the answer is predicted by calculating the probabilities of the first and last tokens of the answer span.  Pre-trained language models such as BERT \cite{devlin-etal-2019-bert}, T5 transformer \cite{raffel2020exploring} and XLNet \cite{yang2019xlnet}, achieve exceptional performance in EQA systems, however at the cost of reliance on large human-annotated supervised datasets. The Stanford Question Answering Dataset (SQuAD)~\cite{rajpurkar-etal-2016-squad} is a widely used dataset for EQA systems.

\citet{lewis-etal-2019-unsupervised}, \citet{fabbri-etal-2020-template}, \citet{li-etal-2020-harvesting}, and \citet{lyu-etal-2021-improving} used randomly sampled passages from Wikipedia, where named entities, or noun chunks, are identified as answers as these tend to be useful for question answering. The questions are then formed in natural language according to the passage and a selected answer phrase.

\begin{figure*}[ht!]
    \centering
    \includegraphics[width=\textwidth-2cm]{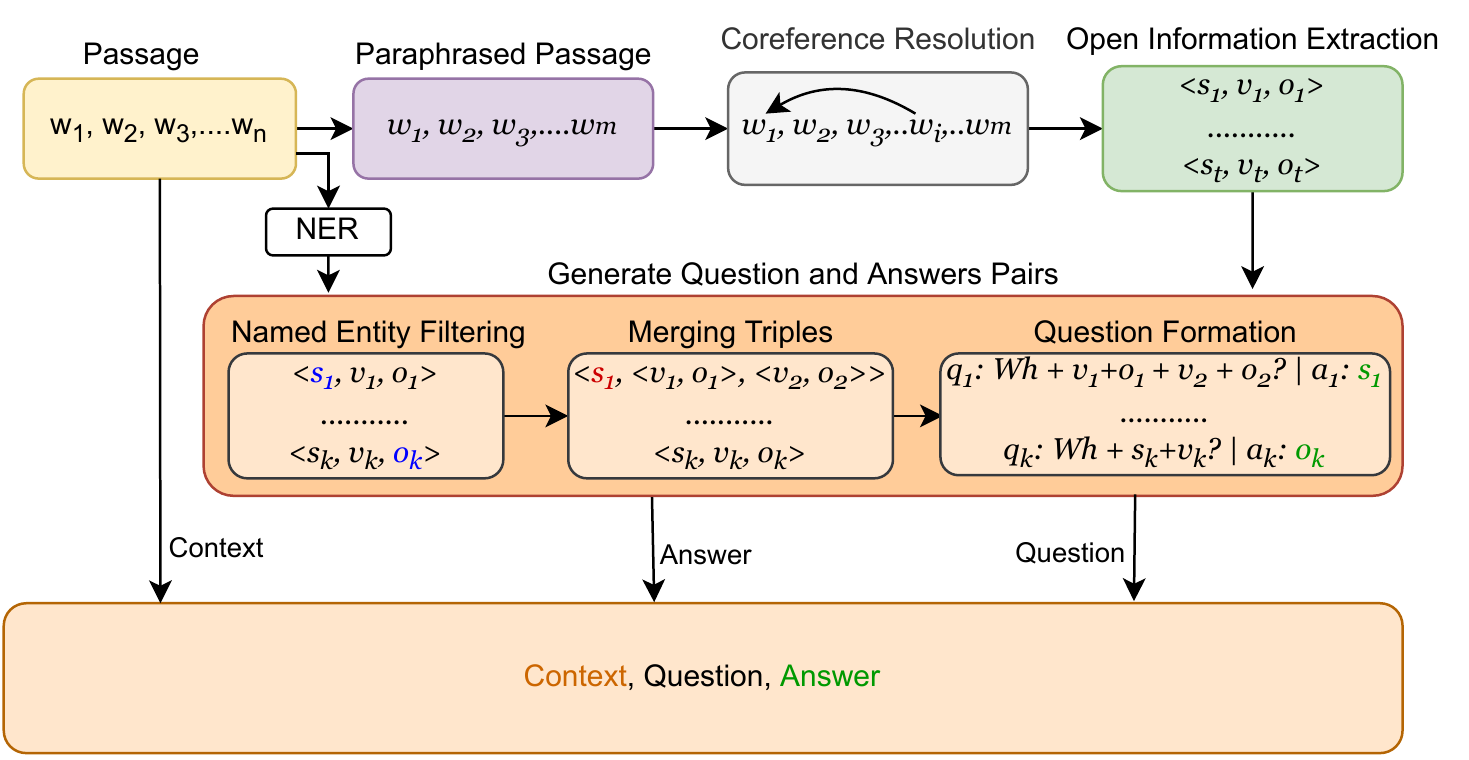}
    \caption{The general pipeline of PIE-QG for question generation using paraphrasing and OpenIE. Note: {\color{blue} Blue}  indicates named entities, {\color{red} red} merged triples with a common subject and {\color{green} green} the selected answers.}
    \label{fig:pipeline}
\end{figure*}

Unsupervised EQA is achieved using the cloze-translation method~\cite{lewis-etal-2019-unsupervised} by forming passage, question-answer triples from a given target corpus. The answers present in the passages are masked to form ``fill in the blanks'' styled questions, so-called cloze questions. The authors translate natural language questions using a neural machine translation (NMT) model trained with different corpora that contain cloze questions and natural question pairs.

Questions generated directly from the passage can only answer simple cloze questions by matching text within the passage, an approach that can not give correct answers for differently phrased questions. 
In an effort to broaden the questions used to train an EQA system,~\citet{fabbri-etal-2020-template} generated questions using a similar sentence taken from a different passage. The actual passage is considered a query and sentences are retrieved using elastic search. The most similar sentence, which contains the answer but excludes the original query passage, and with less than 95\% similarity, to avoid plagiarised sentences, is used to form the question-answer pairs. 
The answer from these sentences is masked and a question in the form of a ``Wh+B+A?'' rule, where ``wh'' (one of what, when, or who) is selected based on the answer-entity type (``B'' is a fragment of the sentence that comes after the answer mask, and ``A'' is the fragment that is present before the answer mask). 

\citet{li-etal-2020-harvesting} uses citations to form a summary of the passage. The cited passage is considered the context, and the sentence where the citation appeared is used for question generation, to avoid lexical overlap. The question generation process involves masking the answer with a cloze mask, where the mask mentions only the type of the answer entity. The dependency tree for the sentence is altered in such a way that the cloze mask is brought to the beginning. The question is then created by replacing the cloze mask with the suitable ``wh'' word, again determined by the type of the answer entities. 

\citet{lyu-etal-2021-improving} perform unsupervised QA by creating a question generation model from text summaries. The model uses dependency trees and semantic role labels extracted from the summary to generate a question. A neural encoder-decoder model is then trained to translate articles to summary-informed questions. The trained model is applied to the actual passages to create questions. However, we consider this method as a transfer learning task rather than unsupervised question generation due to its dependency on a text-summary dataset. Our method compares to \citet{fabbri-etal-2020-template} and \citet{li-etal-2020-harvesting}, avoids the sentence and citation-based retrieval, and minimizes the requirement of having a large corpus to generate question-answer pairs.

\section{Paraphrased Information Extraction for Question Generation}
To overcome the reliance on external reference data sources with a large number of passages, we made use of OpenIE and paraphrased passages for unsupervised synthetic question generation. 
The actual passages are first altered to a paraphrased form and <subject, predicate, object> triples are then extracted from the paraphrased passages. These triples, combined with certain heuristics, form question-answer pairs which are then used alongside the original passage as context to fine-tune the QA model.

The pipeline of our proposed EQA question generation process is illustrated in Figure~\ref{fig:pipeline}. The steps in this pipeline are detailed as follows. 

\textbf{\textit {(i) Paraphrasing:}} Question-answer pairs generated directly from the passage result in inferior QA system performance, as they produce models that have little ability to generalize~\cite{fabbri-etal-2020-template}. Paraphrasing is therefore adopted to alter the passage without changing its actual meaning. The intuition behind this is to create questions from passages that are semantically similar but lexicographically different from the original passage. Paraphrasing question-answer pairs themselves has been shown to cause semantic drift~\cite{pan-etal-2021-unsupervised}. By contrast, in our approach, the passage is paraphrased, rather than question-answer pair. This improves the model's performance. The effect of paraphrasing is discussed in Section~\ref{sec:results}.

\textbf{\textit {(ii) Co-reference resolution:}} As we aim to make use of every sentence in the passage to generate questions, some sentences are ineffective due to the presence of pronouns~\cite{ma2021knowledge}. This problem is solved by implementing co-reference resolution, replacing pronouns in the paraphrased passages with the proper name of the referring noun. 

\textbf{\textit {(iii) Information Extraction:}} OpenIE is applied on paraphrased passages to generate extractions in the form of arguments and relations from natural language text~\cite{mausam2016open}. Given a sentence $w_i$ in the passage, \{$w_1, w_2, w_3,..w_N$\}, OpenIE  generates extractions \{$T_1, T_2, T_3,...T_M$\}, where each extraction is in the form {<subject, predicate, object>}, namely triples. OpenIE is proven to be an efficient solution for 
downstream tasks such as complex question answering~\cite{khot-etal-2017-answering}.

\begin{algorithm}[h!]
\SetAlgoLined
\SetKwInOut{alginput}{Input}
\SetKwInOut{algoutput}{Output}
\alginput{Given a passage $P$ from the corpus }
\algoutput{A list of Question-Answer Pairs}
\BlankLine
    $P'$ = Paraphrase($P$) \\
    $CP$ = Coreference\_Resolution($P'$) \\
    $T$ = Open\_IE($CP$) \\
    {\it named\_entities} = NER($CP$) \\
    $T_{ne}$ = NE\_filter($T$, {\it named\_entities} ) \\
    $T_{IF}$ = IdenticalTriple\_filter($T_{ne}$) \\
    $T_{M}$ = Merge\_Triples($T_{IF}$) \\
    $T_{IF}$ = Remove\_Merged\_Triples($T_{IF}, T_{M}$) \\
    {\it $QA\_Pairs$} $\xleftarrow{}$ $newlist$ \\

    \For{ \mbox{\it $t_n$} in $T_{M}$} {   
        $A$ = Select\_Answer($t_n$) \\
        $Q$ = Wh \\
        \For{ \mbox{\it <subject, relation, object>} in $t_n$} {
            $Q$ = $Q$ + relation + object/subject \\
            $QA\_Pairs \xleftarrow{}  append(\langle Q, A, P\rangle)$ \\
        }
    }      
    
    \For{ \mbox{\it <subject, relation, object>} in $T_{IF}$}
    {   $Q$ = Wh + relation + object? \\
        $A$ = subject \\
        QA\_Pairs $\xleftarrow{}$ append($\langle Q, A, P\rangle$) \\
        \BlankLine
        $Q$ = Wh + subject + relation? \\
        $A$ = object \\
        QA\_Pairs $\xleftarrow{}$ append($\langle Q, A, P\rangle$) \\
    } 
    \Return $QA\_Pairs$
\caption{PIE-QG: Question generation from passages.} \label{alg:question-gen}
\end{algorithm}

\label{sec:question_generation}
\textbf{\textit {(v) Question formation:}} OpenIE extractions produced from a passage are used to form questions as a synthetic training set for QA system fine-tuning.

\textbf{\textit {(vi) Named entity filtering:}} Since triples extracted from a passage have different types of extractions, we select the triples that contain named entities in the answer. In other words, the subject (or object) is selected as an answer only if it is a named entity.

\textbf{\textit {(vii) Eliminating duplicate triples:}} One downside of open information extraction is the presence of duplicate or semantically redundant triples. Generating separate questions from similar or duplicate triples causes  inferior performance in the EQA system model, hence redundant triples are sorted and the longest triple from the sort is selected as the single source for final question generation. 

\textbf{\textit {(viii) Merging triples:}} Questions generated from the triples using the above methods result in simple and easy-to-answer questions. For robust model training, we generate more complex questions from multiple triples by grouping triples with the same subject or object. For instance, if there are two triples of the form \begin{math}\{\langle s_1,r_1, o_1\rangle, \langle s_2,r_2,o_2\rangle\}\end{math} and $s_1=s_2$, we form a question-answer pair with ``Wh + $r_1$ + $o_1$, $r_2$ + $o_2$?'' as the question and $s_1$ (or $s_2$) as the answer.

Each triple extracted from a paraphrased passage can form two questions with either subject or object as an answer. When a subject is selected as an answer, the question is formulated as ``Wh + relation + object?''. Conversely, when an object is selected as the answer, the question generated is of the form ``Wh + subject + relation?''. ``Wh'' is the question word in these formulations and the appropriate form is selected from a list, based on the answer entity type as earlier described.

\section{Experimental Platform}
\label{sec:experiments}
\paragraph{Datasets} The performance of our question generation method is evaluated in terms of Exact-Match (EM) and F-1 score using  existing EQA datasets, namely SQuAD v1.1 \cite{rajpurkar-etal-2016-squad} development set, and NewsQA \cite{trischler2016newsqa}, BioASQ \cite{tsatsaronis2015overview} and DuoRC \cite{saha-etal-2018-duorc} test sets. SQuAD version 1.1 is acquired from the official version\footnote{\href{https://rajpurkar.github.io/SQuAD-explorer/}{https://rajpurkar.github.io/SQuAD-explorer/}}
while the \citet{fisch2019mrqa} published versions of test sets are considered for NewsQA, BioASQ, and DuoRC. A more recent SQuAD v2.0 \cite{rajpurkar-etal-2018-know} is considered unsuitable for our experiments as the synthetic training set does not contain unanswerable questions. 

 \paragraph{Question Generation
} 
We take a relatively small subset of 30,000 passages from the \cite{li-etal-2020-harvesting} sampled Wikipedia dataset for question generation and for  training the model. 
The pseudo-code for the proposed question generation technique is presented in Algorithm~\ref{alg:question-gen}.

Some of the questions resulting from this process can be grammatically incorrect.  We rely on questions posed to the model during inference to be in natural language with correct grammar, we experiment by introducing a grammar correction module in the pipeline to synthesize syntactically accurate questions but later removed this due to its effect discussed in Section~\ref{sec:results}.

\begin{table*}
\centering \small
 \begin{tabular}{lcc|cc|cc|cc}
{}&\multicolumn{2}{c}{SQuAD1.1} & \multicolumn{2}{c}{NewsQA} & \multicolumn{2}{c}{BioASQ}& \multicolumn{2}{c}{DuoRC} \\
\hline
\textbf{PIE-QG Heurisitcs} & \textbf{EM} & \textbf{F-1}  & \textbf{EM} & \textbf{F-1} & \textbf{EM} & \textbf{F-1} & \textbf{EM} & \textbf{F-1} \\
\hline
{Open IE} & {22.8} & {36.5} & {13.0} & {23.9} & {16.6} & {24.3} & {22.0} & {28.5} \\
{+ Paraphrasing} & {37.7} & {53.6} & {19.9} & {32.3} & {20.3} & {31.6} & {32.6} & {40.7}\\
{+ Co-reference Resolution} & {44.2} & {53.4} & {21.1} & {31.8} & {26.5} & {34.7} & {34.8} & {40.4}\\
{+ Named-Entity filter} & {46.6} & {56.5} & {21.7} & {32.5} & \textbf{30.3} & {36.9} & \textbf{36.9} & \textbf{42.4}\\
{+ Filtering Identical Triples} & {47.5} & {57.8} & {21.8} & {32.2} & {30.1} & {37.1} & {35.1} & {41.1} \\
\textbf{+ Merging Triples} & \textbf{48.6} & \textbf{58.7} & {21.8} & \textbf{32.8} & {29.6} & \textbf{37.5} & {34.3} & {40.1}\\
+ {Grammar Correction} & {47.1} & {56.8}  & \textbf{21.9} & {32.3} & {29.1}& {36.5} & {35.2} & {40.7}\\
\hline
\end{tabular}
\caption{Ablation study of the different techniques used in PIE-QG and their subsequent impacts on the EM and F-1 after fine-tuning the BERT-base model. Note: Each step represents an incremental upgrade to the previous step in question generation.}
\label{tab:accents}
\end{table*}

Sourced Wikipedia passages are transformed into paraphrased passages with a pre-trained model\footnote{\url{https://huggingface.co/tuner007/pegasus\_paraphrase}} based on the PEGASUS transformer~\cite{zhang2020pegasus}. Pronouns in the paraphrased passage are replaced with the nouns they refer to. We used neuralcoref\footnote{\url{https://spacy.io/universe/project/neuralcoref}} for this purpose, the spaCy implementation of pre-trained co-referent resolution based on reinforcement learning \cite{clark-manning-2016-deep}.
OpenIE6 is used to extract {<subject, predicate, object>} triples from the pronoun-replaced paraphrased passages. OpenIE6 uses Iterative Grid Labeling and is based on BERT. A spaCy-based named-entity recognition (NER) module~\cite{Honnibal_spaCy_Industrial-strength_Natural_2020} is used to generate a list of named-entities from the passage. Named-entity recognition (NER) is particularly helpful for filtering triples and determining the answer-entity type for appropriate ``wh'' word selection. The simplest version of ``Wh'' word is selected for a particular named entity based on~\citet{fabbri-etal-2020-template}. Questions generated from this process are grammatically corrected using a RoBERTa-based~\cite{liu2019roberta} grammar correction module named ``GECToR''~\cite{omelianchuk-etal-2020-gector}.
 All  models are applied from the above-mentioned sources out-of-the-box, namely  with no domain specific fine-tuning.
 
 \paragraph{QA fine-tuning} We use pre-trained BERT models from \citet{devlin-etal-2019-bert} as the baseline and fine-tune the models for downstream QA system tasks with the generated training data. The generated question, and its context (the actual NL-passage that contains both the question and its answer), are represented as a single sequence, separated by different segment masks and the ``[SEP]'' token. The final linear layer of the model is trained, to identify the start and end spans of the answer, by computing log-likelihood for each token. All  experiments are performed on the uncased version of the BERT-base model with a learning rate of 3e-5, a maximum sequence length of 384, a batch size of 12, a document stride of 128 for 2 epochs, and a check-point at every 500 steps. The best check-point was selected by validating each against 5000 QA pairs randomly sampled from the synthetic training data. We use the Huggingface\footnote{\url{https://huggingface.co}} implementation for input tokenization, model initialization, and training. For comparison with the state-of-the-art EQA models, we also experimented on the BERT-large whole-word masking version with the same training data. All models are trained and validated on a single NVIDIA Tesla A100 GPU.

\section{Results and Discussion}
\label{sec:results}
The effectiveness of the question-answer data generated using the PIE-QG method is measured by training the BERT-base model and evaluating it against existing EQA development and test sets. The Exact Match (EM) and F-1 scores are selected as the metrics to evaluate the effectiveness of each component in the QA models. The initial set of questions is created using OpenIE, where the passage is directly used to form triples and generate questions as described in Section~\ref{sec:question_generation}. The intuition behind using OpenIE is to generate multiple questions from a single passage. However, as previously described, such a simple-minded approach suffers from having pronouns as answers, ungrammatical questions, and high degrees lexical similarity between passage and question, making most extracted triples suitable  for word matching only.

\begin{table*}[ht!]
\centering \small
\begin{tabular}{lll|ll|ll|ll|p{0.55in}ll}
{}&\multicolumn{2}{c}{SQuAD1.1}  & \multicolumn{2}{c}{NewsQA} & \multicolumn{2}{c}{BioASQ}& \multicolumn{2}{c}{DuoRC} &\\
\hline
\textbf{Fine-tuning Models} & \textbf{EM} & \textbf{F-1}  & \textbf{EM} & \textbf{F-1} & \textbf{EM} & \textbf{F-1} & \textbf{EM} & \textbf{F-1} & \textbf{\#Training Contexts}\\
\hline

\textit{BERT-base} \\
Sentence Retrieval \cite{fabbri-etal-2020-template} & {46.1}$\dagger$ & {56.8}$\dagger$  & {20.1} & {31.1} & {29.4} & \textbf{38.1} & {28.8}& {35.0}& {45K} \\
\textbf{PIE-QG (Ours)} & \textbf{48.6} & \textbf{58.7} & \textbf{21.8}& \textbf{32.5} & \textbf{29.6} & {37.5} & \textbf{34.3} & \textbf{40.1} & \textbf{20-28K}\\
\hline
\textit{BERT-large} \\
Cloze Translation \cite{lewis-etal-2019-unsupervised} $\dagger$ & {45.4} & {55.6}& {19.6} &	{28.5} & {18.9} & {27.0} & {26.0} & {32.6} & 782K \\
RefQA \cite{li-etal-2020-harvesting} & {57.1} $\dagger$ & {66.8} $\dagger$ & {27.6} & {41.0} & {42.0}& {54.9}& {41.6} & {49.7} & 178K \\
+ Iterative Data Refinement & \textbf{62.5} $\dagger$ & {\bf 72.6} $\dagger$ & \textbf {32.1} & \textbf{45.1} & \textbf{44.1}	& \textbf{57.4} & \textbf{45.7} & \textbf{54.2} &240K \\
\textbf{PIE-QG (Ours)} & {61.2} & \textbf{72.6} & {29.7}	& { 44.1} & {43.6} & {55.1} &  { 44.6} & {52.9} & \textbf{20-28K}\\
\hline
\end{tabular}

\caption{Comparison of PIE-QG with state-of-the-art unsupervised QA models. Note: Iterative refinement achieves the best performance through structural analysis of the corpus via citation and intra-document links, a model that requires $\times8$ as many contexts as the PIE-QG model we propose.`$\dagger$' indicates results taken from the existing literature, and all other figures  are evaluated with published synthetic training data (or) pre-trained models. ``\#Training Contexts'' are measured based on respective published synthetic datasets. Each model uses the same synthetic training data sourced from Wikipedia for fine-tuning and is evaluated against the standard EQA datasets.}
\label{tab:sota}
\end{table*}

\paragraph{Effect of Paraphrasing}
Using paraphrased passages for question generation avoids lexical overlaps with the passage and improves model performance. Ten different paraphrases are generated for each sentence in the passage using the PEGASUS~\cite{zhang2020pegasus} paraphrasing generation model. Jensen-Shannon Divergence (JSD) is calculated for each paraphrase against the original sentence. JSD calculates a divergence score based on the word distributions between two sentences, a higher value for JSD accounts for a more different sentence, while a lower value JSD score represents higher lexical overlap. In our PIE-QG pipeline, sentences with the highest JSD values are selected for question generation to make the question syntactically different. {Paraphrasing} has a strong positive effect on the model, improving the EM \& F-1 score by at least 4\% and 7\% respectively on all evaluation sets.

\paragraph{Effect of Co-reference Resolution}
The presence of pronouns in passages results in meaningless question-answer pairs. For instance, \textit{``Vaso Sepashvili (; born 17 December 1969) is a retired Georgian professional footballer. He made his professional debut in the Soviet Second League B in 1990 for FC Aktyubinets Aktyubinsk''} is the passage. This produces a triple \textit{``<He, made, his professional debut in the Soviet Second League B in 1990 for FC Aktyubinets Aktyubinsk>''}. While the relation and object form a question ``Who made his professional debut in the Soviet Second League B in 1990 for FC Aktyubinets Aktyubinsk?'' with the subject ``He'' selected as the answer. The best answer for this question is found co-referenced in the previous sentence where the pronoun ``He'' refers to ``Vaso Sepashvili''. To address this we  alter the passage with co-reference resolution to replace all pronouns with the referring proper noun. The above sentence is changed in such a way that the extracted triple becomes \textit{``<Vaso Sepashvili, made, his professional debut in the Soviet Second League B in 1990 for FC Aktyubinets Aktyubinsk>''} and the ideal answer is selected. Pronouns were replaced with their referring nouns using this method to generate meaningful questions while the original passage is retained for training the QA model.  In this way, {co-referent resolution} has a positive impact on the model performance increasing the EM by 2\%-6\% across all the sets.

\paragraph{Named-Entity Filtering}
As triples are the direct source of training questions, the quality of triples leads to better training questions for the PIE-QG model. In general, OpenIE6 returns all possible triples from a sentence, but selecting suitable triples, to generate better question-answer pairs, becomes important. To assist in identifying the best set of triples, we filter triples that do not contain named entities. We use Named Entity Recognition (NER) to extract all named entities from the passage. To become a candidate to be selected for the question generation process, either the subject or object from the triple must contain at least one named entity.  This NER filtering method is beneficial to the model, it eliminates many impractical question-answer pairs from the training set and improves the overall Exact Match (EM) and F-1 score by 2\% except for NewsQA. 

\paragraph{Effect of Filtering Identical Triples}
Semantically similar triples are formed using OpenIE6 with a high degree of lexical overlap. Constructing questions from these triples causes question duplication and has the potential to deteriorate model performance and even result in  over-fitting. To filter similar or duplicate triples, each triple is verified with other triples extracted from the passage to discover lexical overlaps between them. If a triple formed as a sentence is a sub-string of another, the shorter is removed from the training set to avoid the production of redundant questions. From Figure~\ref{fig:qa-sample}, triples such as \textit{<the deals, could violate, EU antitrust laws>} and
\textit{<The European Commission, is worried, that the deals could violate EU antitrust laws>} convey the same meaning with a high degree of lexical overlap, hence the former is removed. {\it Filtering identical triples} in this way has a small but favorable impact on the model as shown in the ablation summary in Table~\ref{tab:accents}.

\paragraph{Effect of Merging Triples}
A subject (or object) in a passage can exhibit relations to multiple objects (or subjects). Triples with common subjects are merged to form complex questions such that QA model can understand complex relationships. {Merging triples} has a small but positive effect on the model performance improving EM by 1.2\% and F-1 by 0.9\% as shown in Table~\ref{tab:accents}. 

\paragraph{Effect of Grammar Correction}
Questions generated from the above process often contain grammatical errors which can negatively impact model performances. We experimented with ``GECToR'' \footnote{\href{https://github.com/grammarly/gector}{https://github.com/grammarly/gector}},  a grammar correction module that tags and corrects input questions with grammar errors. For instance, the question ``What is is worried that the deals could violate EU antitrust laws?'' is formulated. The repeat occurrence of the verb ``is'' is an obvious error. The grammar correction module alters the question where the final question is formulated correctly as ``What is worried that the deals could violate EU antitrust laws?''. Based on heuristics presented in Table~\ref{tab:accents}, all incremental upgrades until ``Merging Triples'' improve the model performance, but {Grammar correction} does not and is hence removed from the pipeline. 

\begin{figure}[h!]
    \centering
    \includegraphics[width=0.5\textwidth]{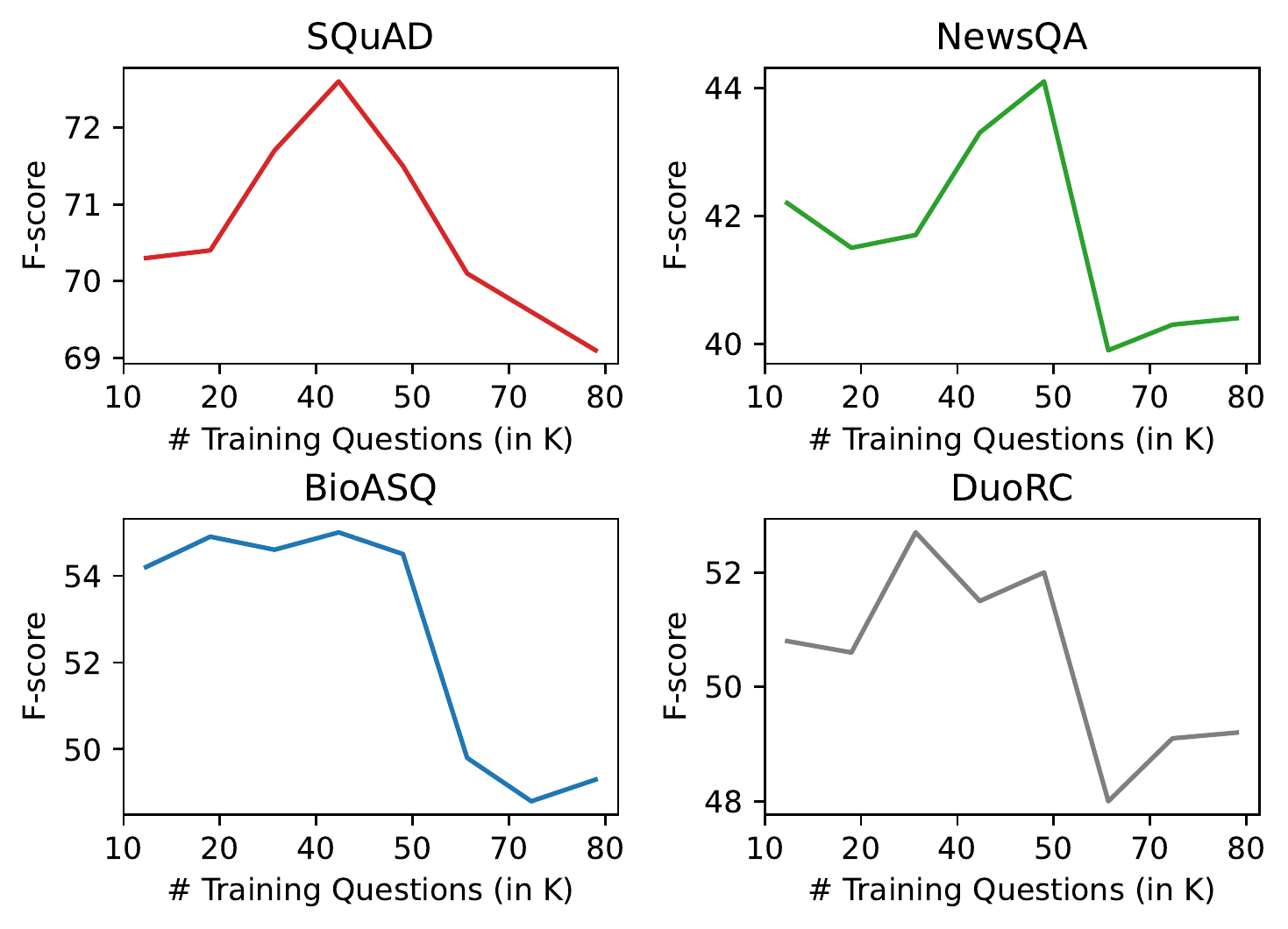}
    \caption{Evaluation of the  PIE-QG model F-score for different datasets against the number of questions in the training set using the BERT-large model, the optimal number for each dataset is in the range 30-50K.}
    \label{fig:training_questions}
\end{figure}

\paragraph{Effect of Training Data Size} Experiments were conducted to measure the EM and F-score at different synthetic data sizes to identify the optimal number of training questions. Figure~\ref{fig:training_questions} presents the results of these experiments and  reveals that PIE-QG achieves peak performance between 30K-50K training questions using BERT-large model and begin to over-fit beyond that number. The same effect is also observed in \cite{fabbri-etal-2020-template}. 
The method to determine the optimal number of training questions is to split the generated question-answer pairs into blocks each of 10K. These are then split into training and validation sets. At fixed points of 500 training steps, the validation set is measured against the QA model. This incrementally informs the process of when the model optimizes against the number of question-answer pairs used to train it. It is observed, shown in Figure~\ref{fig:training_questions}, that this occurs for each of the datasets in the range 30-50K. Increasing the number of template-styled training questions negatively affects the evaluation performance after a certain point because of memorisation of synthetic data patterns. 

\paragraph{Comparison with the State-of-the-Art}
\citet{fabbri-etal-2020-template} use a BERT-base model as the backbone for their experiments while \citet{lewis-etal-2019-unsupervised} and \citet{li-etal-2020-harvesting} employed the BERT-large whole word masking pre-trained model. Questions generated from the PIE-QG model performed better than the information retrieval-based method presented by \citet{fabbri-etal-2020-template} and produced an absolute improvement of 2.5\% on EM and 1.9\% on F-1 on the SQuAD 1.1 development set. Comparing BERT-large models, the PIE-QG model outperforms citation retrieval-based RefQA, a method  that involves dependency tree reconstruction. However, RefQA, which includes a refinement technique, achieves the best performance, achieving 1-2.5\% higher F-1 score than that of PIE-QG, but at the cost of using $8\times$ more passages and $10\times$ more training questions. Also, refinements in RefQA are performed on the training data through iterative cross-validations on the SQuAD 1.1 development set, whereas the PIE-QG model does not involve such a process. The number of passages and questions used by each method are presented in detail in Table \ref{tab:num_passages}. PIE-QG outperforms retrieval-based question generation on every dataset and produces comparable performance with RefQA with $8\times$ fewer passages.

To summarise, the experimental results demonstrate the advantages of the PIE-QG method;  
\begin{enumerate}
    \item Paraphrasing the original passage eliminates the need of using external knowledge sources to avoid lexical overlap; 
    \item Multiple questions generated using OpenIE with our proposed method minimizes the requirement of a large corpus without having to sacrifice the performance.
\end{enumerate}

\begin{table}
    \centering \small
    \begin{tabular}{p{0.22\textwidth}p{0.1\textwidth}p{0.08\textwidth}}
        \textbf{System} & \textbf{\#Contexts} & \textbf{\#Questions} \\
        \hline
        \citet{fabbri-etal-2020-template} &  45K & 50K\\ 
        RefQA \citet{li-etal-2020-harvesting} &  178K & 300K\\ 
        + IDR & 240K & 480K \\
        PIE-QG & 20-28K & 30-50K \\
        \hline
    \end{tabular}
    \caption{Comparison of statistics of the synthetic training data generated by existing unsupervised question generation methods with PIE-QG.}
    \label{tab:num_passages}
\end{table}

\section{Limitations}
\label{sec:limitations}
The downside of the PIE-QG unsupervised question generation pipeline is the use of external modules like paraphrasing, OpenIE, and NER, which may not exist in languages other than English. The quality of question-answer pairs generated to train the QA model is therefore dependent on the performance of these modules on the selected corpus. It is however anticipated that PIE-QG will perform similarly well on any English language corpus. It is future work to apply these modules within the PIE-QG pipeline to other languages where comparable language-specific models can be sourced and  performance outcomes analyzed.

\begin{table*}
    \centering

    \begin{tabular} {p{0.03\linewidth} | p{0.91\linewidth}}
     \hline
        \textbf{P} & Georgia Tech undergraduate programs continue to excel, and I’m pleased that we’ve been able to maintain this measure of excellence for so long, ” said Interim President and Provost Gary Schuster. \\
        \textbf{Q} & Who was said that he was pleased that Georgia Tech undergraduate programs continued to excel? \\
        \textbf{A} &\textbf{ Gary Schuster} \\
        \hline
        
        \textbf{P} & ``We're very upset, very angry,'' said Raphael Felli, 35, a 
        U.S.- based attorney and son of executed Colonel Roger Felli, who was foreign minister in the Acheampong administration. \\
        \textbf{Q} & Who was is an attorney based in the U.S., is the son of executed Colonel Roger Felli? \\
        \textbf{A} & \textbf{Raphael Felli} \\
        \hline
        
        \textbf{P} & Liberty and Tyranny Sells a Million. Politics Radio host Mark R. Levin’s bestselling Liberty and Tyranny : A Conservative Manifesto has sold one million copies, according to publisher Threshold Editions.....Published on March 24, 2009, Liberty debuted at \# 1 on the New York Times bestseller list. \\
        \textbf{Q} & What sell a million, made it to the New York Times bestsellers list? \\
        \textbf{A} & \textbf{Liberty} \\
        \hline
    \end{tabular}
    \caption{
    Example synthetic question-answer pairs generated using PIE-QG. Note: \textbf{P} represents the passage extracted from a document. \textbf{Q} and \textbf{A} are the generated question and the selected answer from the passage, respectively.}
    
    \label{tab:example_qa_pairs}
\end{table*}

\section{Conclusion}
\label{sec:conclusion}
With no reliance on any external reference corpora, the PIE-QG model uses paraphrasing and Open Information Extraction (OpenIE) to generate synthetic training questions for fine-tuning the language model in a QA system based on BERT. Triples in the form of <subject, predicate, object> are extracted from paraphrased passages, and questions are formed with subjects (or objects) as answers. Pronoun co-referents are resolved and where possible, triples are merged, and duplicate and highly similar triples are removed. Furthermore, triples that do not contain named entities are eliminated. 
The PIE-QG pipeline results in a high-quality question-answer training set that informs the QA model. Using the PIE-QG pipeline  results in a QA model that achieves performance comparable to the state-of-the-art performance using significantly fewer passages. It is only narrowly outperformed by \mbox{RefQA}, an approach that uses iterative data refinement, and therefore relies on the citation structure of corpora and $\times10$ more training questions.

\bibliographystyle{acl_natbib}
\bibliography{anthology,custom}




\end{document}